\documentclass[11pt]{article}

\usepackage[final]{acl}

\usepackage{times}
\usepackage{latexsym}

\usepackage[T1]{fontenc}

\usepackage[utf8]{inputenc}

\usepackage{microtype}

\usepackage{inconsolata}

\usepackage{graphicx}
\usepackage{booktabs}
\usepackage[whole]{bxcjkjatype} 
\usepackage{url}

\usepackage{multicol}
\usepackage{multirow}
\usepackage{booktabs}
\usepackage{amsmath,amssymb}
\usepackage{bm}
\usepackage{bbm}
\usepackage{tabularx}
\usepackage{tablefootnote}
\usepackage{float}
\usepackage{xspace}
\usepackage{CJKutf8}
\usepackage{listings}
\usepackage{enumitem}
\usepackage[subrefformat=parens]{subcaption}
\usepackage{fontawesome5}
\usepackage{tcolorbox}
\usepackage{adjustbox}
\usepackage{array}
\usepackage{fontawesome5}
\usepackage{subcaption}
\newcommand{\update}[1]{#1}

\title{Edit-level Majority Voting Mitigates Over-Correction \\in LLM-based Grammatical Error Correction}

\author{
  \textbf{Takumi Goto}, \
  \textbf{Yusuke Sakai}, \
  \textbf{Taro Watanabe}
\\
  Nara Institute of Science and Technology (NAIST)
\\
  \texttt{\{goto.takumi.gv7, sakai.yusuke.sr9, taro\}@is.naist.jp}
}

\begin{document}
\maketitle
\begin{abstract}
Grammatical error correction using large language models often suffers from the over-correction issue. To mitigate this, we propose a training-free inference method that performs edit-level majority voting over multiple candidates generated by a single model, without requiring model modifications or additional training. Across nine benchmarks covering English, Czech, German, Ukrainian, Korean, Hindi, and Romanian, the proposed method outperforms both greedy and MBR decoding in most cases. Moreover, it yields stable correction quality regardless of the instruction prompts used. 
\update{We release two repository supporting GEC datasets loading\footnote{\url{https://github.com/gotutiyan/gec-datasets}} and LLM inference\footnote{\url{https://github.com/gotutiyan/gec-llm/}}.}
\end{abstract}

\section{Introduction}

In the era of Large Language Models (LLMs), users can employ LLMs for Grammatical Error Correction (GEC) via prompting and few-shot examples. However, prior work has shown that while LLMs perform well on fluency editing, they struggle to adhere to minimal editing~\cite{katinskaia-yangarber-2024-gpt,omelianchuk-etal-2024-pillars}, a limitation known as the \textit{over-correction} problem.
Over-correction arises when LLMs modify grammatically correct text to improve fluency or generate irrelevant outputs such as \textit{``Okay, this is the corrected sentence:...''}. This issue persists even with carefully designed prompts and few-shot examples.
Over-correction not only degrades user experience but also affects evaluation reliability. Since GEC performance is typically measured using the $F_{0.5}$ score, which prioritizes precision~\cite{ng-etal-2014-conll,bryant-etal-2017-automatic,gong-etal-2022-revisiting}, \update{models that frequently exhibit over-correction contradict such metric's requirements.}
GEC models are expected to modify only grammatically incorrect spans, a constraint emphasized in the ``minimal edit'' setting. Prior work has explored approaches such as copy mechanisms in sequence-to-sequence models~\cite{zhao-etal-2019-improving} and sequence labeling-based methods~\cite{awasthi-etal-2019-parallel, omelianchuk-etal-2020-gector} to enforce this behavior.
Although recent studies for LLMs mitigate over-correction via fine-tuning~\cite{liang-etal-2025-edit,staruch-etal-2025-adapting}, such approaches introduce significant computational overhead and require maintaining fine-tuned models for each base model.

\begin{figure}[t]
\centering
    \includegraphics[width=0.99\linewidth]{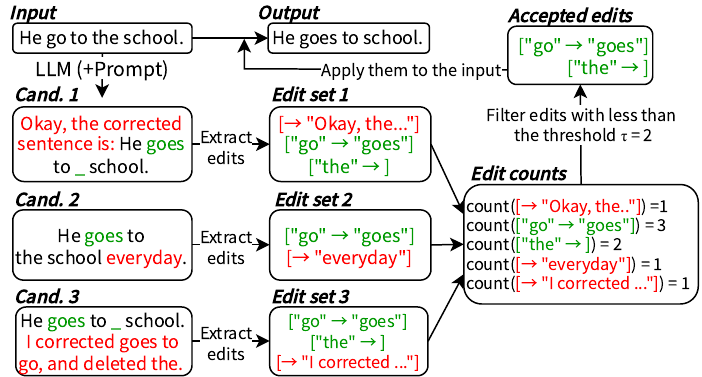}
    \caption{Overview of the edit-level majority voting process using three outputs generated from a single LLM. Extraneous text and over-corrections (e.g., [→ “everyday”]) shown in red are removed, while only the corrections indicated in green are adopted.}
    \label{fig:overview}
\end{figure}

To address over-correction without additional training, we propose a training-free inference method that performs edit-level majority voting over multiple candidates generated by a single LLM. The key intuition is that \emph{valid corrections are more likely to appear consistently across sampled outputs than spurious ones.}
As illustrated in Figure~\ref{fig:overview}, we sample multiple candidates and retain only those edits that occur in at least a predefined number of samples. Although closely related to majority voting ensembles~\cite{tarnavskyi-etal-2022-ensembling}, our approach differs in that it leverages multiple samples from a single model rather than multiple models. Moreover, the method can be interpreted as capturing edit-level self-consistency~\cite{DBLP:conf/iclr/0002WSLCNCZ23}, i.e., edits that are repeatedly generated across independent samples are unlikely to arise from stochastic variation and instead reflect decisions with higher model confidence~\cite{deguchi-etal-2024-mbrs, eikema-aziz-2020-map, DBLP:conf/iclr/0002WSLCNCZ23}.

We conduct a comprehensive evaluation in a 4-shot setting across several languages: English, Czech, German, Ukrainian, Korean, Hindi, and Romanian. A comparison between greedy inference and the proposed method reveals that $F_{0.5}$ and GLEU~\cite{napoles-etal-2015-ground,napoles2016gleutuning} scores improve across most datasets. Specifically, on the English CWEB-G~\cite{flachs-etal-2020-grammatical} dataset, $F_{0.5}$ increased by a maximum of over 10 points.%

Based on our analysis, we report the following three findings. \underline{First}, there is a trade-off between correction performance and computational cost, as the proposed method requires generating multiple candidates, but eight candidates are sufficient. \underline{Second}, we present a case study demonstrating that higher thresholds retain only higher-confidence edits, effectively mitigating over-correction. \underline{Finally}, we report achieving a high mean and low variance across ten different prompt templates. This result indicates that, by removing unnecessary text and retaining only high-confidence edits, the proposed method can achieve high scores across templates.

\section{Background and Related Work}\label{sec:background}

The mainstream approach for applying LLMs to GEC is primarily fine-tuning them on GEC training data. This includes simple instruction tuning~\cite{omelianchuk-etal-2024-pillars}, training to generate sequences of edit operations instead of corrected sentences~\cite{kaneko-okazaki-2023-reducing}, Edit-level Preference Optimization (EPO)~\cite{liang-etal-2025-edit}, which captures reinforcement learning rewards at the edit level, and training techniques that optimize data usage strategies~\cite{staruch-etal-2025-adapting}. While these methods are known to achieve high performance on English datasets, it is not always possible to obtain sufficient training data across domains or languages, making additional training difficult. Furthermore, even with lightweight tuning methods such as LoRA~\cite{DBLP:conf/iclr/HuSWALWWC22}, the trained weights must be managed for each base model, leading to increased complexity when experimenting with a wide variety of models. The proposed method overcomes these drawbacks by not requiring fine-tuning.

Correction methods that do not require fine-tuning are primarily based on prompt engineering. \citet{loem-etal-2023-exploring} demonstrated that prompt content impacts correction performance~\cite{davis-etal-2024-prompting}. For instance, providing instructions unrelated to GEC leads to a degradation in correction performance. Furthermore, \citet{davis-etal-2024-prompting} explored prompts that yield higher correction performance, while~\citet{katinskaia-yangarber-2024-gpt} evaluated the zero-shot correction performance of LLMs across multiple languages. \citet{tang-etal-2024-ungrammatical} proposed a method for appropriately selecting few-shot examples and showed that it improves correction performance.
\update{These existing studies discuss strategies for the input to LLMs. In contrast, our approach discusses strategies on the output side.}

\section{Proposed Methods}\label{sec:voting}

We propose a training-free inference method that applies edit-level majority voting to adapt LLMs to minimal editing. 
By leveraging self-consistency in LLM-based GEC, the method mitigates over-correction and improves overall performance.

\subsection{Edit-level Majority Voting} 
Edit-level majority voting is a representative ensemble method in GEC~\cite{tarnavskyi-etal-2022-ensembling, omelianchuk-etal-2024-pillars}. The inputs consist of an erroneous sentence $S$ and $k$ correction candidates $\{H_1, H_2, \dots, H_k\}$. Generally, these correction candidates are obtained by outputting a single corrected sentence from each of $k$ different GEC models. In edit-level majority voting, each set of edits $\{\bm{E}_1, \bm{E}_2, \dots, \bm{E}_k\}$ is extracted using an automatic edit extraction tool such as ERRANT~\cite{felice-etal-2016-automatic, bryant-etal-2017-automatic}. Here, $\bm{E}_i$ represents the set of edits required to transform $S$ into $H_i$.
Given the $\{\bm{E}_1, \bm{E}_2, \dots, \bm{E}_k\}$, we take the union of the $k$ edit sets and, for each edit in this union, count how many times it appears in the $k$ edit sets. Formally, for each edit $e \in \bm{I}$ in the union $\bm{I} = \bm{E}_1 \cup \bm{E}_2 \cup \dots \cup \bm{E}_k$, the count $\mathrm{count}(e)$ is calculated using the following equation.
\begin{equation}
    \mathrm{count}(e) = \sum_{i=1}^{k} \delta(e, \bm{E}_i)
\end{equation}
\begin{equation}
    \delta(e, \bm{E}) = \left\{
\begin{array}{ll}
1 & \text{if} \; e \in \bm{E} \\
0 & \text{otherwise},
\end{array}
\right.
\end{equation}
e.g., $\mathrm{count}($[``\textit{go}'' → ``\textit{goes}'']$) = 3$ in Figure~\ref{fig:overview}.
After counting each edit, we extract only the subset of edits $\bm{I}_{\mathrm{accepted}}$ whose counts are equal to or greater than a threshold $\tau\; (1 \leq \tau \leq k, \tau \in \mathbb{N})$:
\begin{equation}\label{eq:filter}
    \bm{I}_{\mathrm{accepted}} = \{e \mid e \in \bm{I}, \tau \leq \mathrm{count}(e)\}.
\end{equation}
The final corrected sentence is obtained by applying the edits in $\bm{I}_{\mathrm{accepted}}$ to the original erroneous sentence $S$. Note that the number of candidates $k$ and the threshold $\tau$ are hyperparameters.

\subsection{Majority Voting as Self-consistency}

We assume that, when multiple candidates are sampled from an LLM, edits corresponding to minimal edits occur more consistently than those for fluency edits. For instance, corrections for grammatical errors, e.g., missing articles, are relatively constrained, whereas fluency-oriented rewrites tend to be more diverse. Therefore, minimal editing can be encouraged by retaining only high-frequency edits across sampled candidates. This perspective extends the notion of \textit{self-consistency}~\cite{DBLP:conf/iclr/0002WSLCNCZ23} to the edit level.
Given an input sequence $S$ and a GEC prompt, we sample $k$ candidates from a single LLM using nucleus (top-$p$) sampling~\cite{DBLP:conf/iclr/HoltzmanBDFC20}. We then aggregate the sampled candidates via majority voting and retain only high-frequency edits, which can be interpreted as high-confidence corrections. This method is training-free, does not rely on internal model representations, and is thus applicable to black-box LLMs.

Compared to prior work on majority voting ensembles~\cite{tarnavskyi-etal-2022-ensembling, omelianchuk-etal-2024-pillars}, our approach differs in that it leverages multiple samples from a single model rather than multiple models. Moreover, alternative ensemble methods such as ESC~\cite{qorib-etal-2022-frustratingly} and GRECO~\cite{qorib-ng-2023-system}, while applicable, do not explicitly leverage edit-level self-consistency and often require additional training.

\section{Experiments}\label{sec:exp}

\subsection{Datasets and Metrics}\label{subsec:data-metric}
\begin{table}[t]
    \centering
    \small
    \setlength{\tabcolsep}{4pt}
    \renewcommand{\arraystretch}{1.15}
    \begin{tabular}{@{}llrrl@{}}
    \toprule
    Dataset  & Lang. & \# Dev. & \# Test & Metric \\
    \midrule
BEA-2019 & en & 4,384 & 4,477 & ERRANT \\
CWEB-G & en & 3,867 & 3,981 & ERRANT \\
JFLEG & en & 754 & 747 & ERRANT, GLEU \\
AKCES-GEC & cs & 2,485 & 2,676 & JP-ERRANT (cs) \\
Falko-Merlin & de & 2,503 & 2,337 & JP-ERRANT (de) \\
UNLP-2023 & uk & 1,509 & 1,353 & JP-ERRANT (uk) \\
Kor-learner & ko & 4,264 & 4,265 & JP-ERRANT (ko) \\
Hi-GEC & hi & 976 & 1,465 & GLEU \\
RONACC & ro & 451 & 451 & GLEU \\
    \bottomrule
    \end{tabular}
    \caption{Datasets and metrics used in our experiments.}
    \label{tab:datasets}
\end{table}

To evaluate the effectiveness of the proposed inference method, we conduct experiments on multilingual datasets. Table~\ref{tab:datasets} provides a summary of the datasets and metrics used in our experiments. 

For English, we conduct experiments using three types of benchmarks: CWEB~\cite{flachs-etal-2020-grammatical}, BEA-2019~\cite{yannakoudakis2018developing,bryant-etal-2019-bea}, and JFLEG~\cite{napoles-etal-2017-jfleg}. All of these are English datasets comprising both development and test sets, though they differ in the frequency of required error corrections. CWEB-G contains the fewest errors in these datasets, making it the domain where over-correction is most problematic. While BEA-2019 contains more errors than CWEB-G, it still requires minimal editing. JFLEG permits not only grammatical error corrections but also corrections to make a sentence fluent, thus requiring the highest number of corrections. Given that the proposed method is intended to mitigate over-correction, we expect its effectiveness to follow the order of CWEB-G $>$ BEA-2019 $>$ JFLEG. We use ERRANT $F_{0.5}$ as the common evaluation metric for all datasets, and additionally use GLEU for JFLEG.

For other languages, we use AKCES-GEC~\cite{naplava-straka-2019-grammatical} for Czech, Falko-Merlin~\cite{boyd-2018-using} for German, UNLP-2023~\cite{syvokon-romanyshyn-2023-unlp} for Ukrainian, Kor-learner~\cite{yoon-etal-2023-towards} for Korean, Hi-GEC~\cite{sharma-bhattacharyya-2025-hi} for Hindi, and RONACC~\cite{9288338} for Romanian. Regarding evaluation metrics, we use $F_{0.5}$ of JP-ERRANT~\cite{wang-etal-2025-refined,qiu-etal-2025-multilingual} for AKCES-GEC, Falko-Merlin, and Kor-learner; and GLEU for Hi-GEC and RONACC. JP-ERRANT is a Stanza~\cite{qi-etal-2020-stanza} based edit extraction tool and explicitly supports English, Czech, German, Ukrainian, Korean, and Chinese. For UNLP2023, we use the official CodaLab platform because its references are not publicly available. For both ERRANT and JP-ERRANT, we use \textsc{gec-metrics}~\cite{goto-etal-2025-gec} for the implementation of score calculation.

\subsection{Majority Voting Settings}\label{subsec:method-settings}
In Equation~\ref{eq:filter}, we fix the number of generated candidates $k$ to $k=8$. The threshold $\tau$ is determined for each dataset using the development data; we select the threshold that maximizes the metric shown in Table~\ref{tab:datasets} and use it for inference on the test data. For JFLEG, we use GLEU to determine the threshold. For edit extraction tools, we use ERRANT for English datasets and JP-ERRANT for datasets in other languages. Although JP-ERRANT does not explicitly support Hindi and Romanian, it can be used as a tentative edit extraction tool for these languages by employing its multilingual setting. In Nucleus Sampling, we always use $p=1.0$ for top-$p$, 1.0 for temperature.

\subsection{Models and Inference}\label{subsec:model-inference}

To evaluate the utility across multiple models, we use \texttt{Llama-3.1-8B-Instruct}~\cite{grattafiori2024llama3herdmodels}, and \texttt{Qwen3-8B}~\cite{yang2025qwen3technicalreport} for all of datasets. Additionally, we report results of \texttt{gemma-2-9b-it}~\cite{gemmateam2024gemma2improvingopen} for English datasets and \texttt{gemma-3-12b-it}~\cite{gemmateam2025gemma3technicalreport} for other languages, given the scores on the development set. In this paper, we always perform inference in a 4-shot setting. For each dataset, a fixed set of 4-shot samples is used for all inputs. For the English datasets, we use the 4-shot samples employed by \citet{davis-etal-2024-prompting}. For the datasets in other languages, we sort the corresponding training data by length and select the middle four instances as the 4-shot samples. This selection method is adopted for its high reproducibility and to avoid selecting extremely short or long samples as few-shot examples. While some prior studies employ methods to retrieve appropriate few-shot samples for each input sentence using similarity measures such as BM25 or BERT~\cite{devlin-etal-2019-bert} embeddings~\cite{tang-etal-2024-ungrammatical}, we do not adopt such approaches in this paper as they may not be applicable to all languages. For instance, in low-resource scenarios, it may not be possible to secure a sufficient pool of GEC data for retrieval. In contrast, fixed 4-shot samples can be manually created in practice.  We use the vLLM library
~\cite{DBLP:conf/sosp/KwonLZ0ZY0ZS23} for the implementation of inference.

Regarding the instruction template, we use the \texttt{TOOL} template from \citet{davis-etal-2024-prompting}. Note that, to minimize over-correction as much as possible, we append a phrase to the end of the prompt specifying that only the corrected sentence should be output. For non-English datasets, we utilize the template adopted in MultiGEC. This template is also based on the same \texttt{TOOL} template, but it differs in that it explicitly specifies the language. The actual prompts are shown in Appendix~\ref{sec:appen:prompts}.

\subsection{Baselines}

\paragraph{Greedy: Greedy decoding}
Greedy decoding always selects the token with the highest probability from the estimated probability distribution. In nucleus sampling, this is equivalent to always setting the temperature to 0.

\paragraph{MBR: MBR Decoding}
Minimum Bayes Risk (MBR) decoding~\cite{GOEL2000115} is a decoding approach that generates multiple hypotheses and selects the optimal one based on the consensus among them. In a typical formulation~\cite{eikema-aziz-2020-map,deguchi-etal-2024-mbrs}, it performs reference-based evaluation by assuming that the hypotheses themselves also serve as approximated reference sentences, taking the hypothesis with the highest score as the final output $H$:
\begin{equation}\label{eq:mbr}
H = \underset{\hat{H} \in \{H_1, H_2, \dots, H_{k}\}}{\mathrm{argmax}} \frac{1}{k} \sum_{i=1}^{k} u(\hat{H}, H_i).
\end{equation}
$u(\cdot)$ is a utility function to evaluate the quality of each hypothesis $\hat{H}$, i.e., a reference-based metric. By following~\citet{raina-gales-2023-minimum}, we use the evaluation metric as is for the utility function. For example, we use ERRANT $F_{0.5}$ for BEA-2019, and JP-ERRANT $F_{0.5}$ for AKCES-GEC as $u(\cdot)$.
We include MBR decoding as a baseline because it is similar to the proposed method in that it relies on multiple hypotheses. Note that the two methods differ in how to make the final hypothesis: while MBR decoding selects the best one from the candidates, the proposed method integrates them to generate a new candidate.

\paragraph{GEC models other than LLMs}
To provide a comparison with existing non-LLM correction models, we also include GECToR~\cite{omelianchuk-etal-2020-gector} and T5~\cite{2020t5,rothe-etal-2021-simple} based on our reproduction experiments. GECToR performs corrections using sequence labeling, while T5 is based on a sequence-to-sequence model. For further details on the experimental setup, please refer to Appendix~\ref{appen:exp-setting}. We also report the current state-of-the-art performance to discuss the gap between our performance and it.

\subsection{Making Implementation Public}\label{subsec:imple}
\begin{lstlisting}[float=t, caption = {A Python implementation example using our implementation, which covers data loading, correction. The dataset is downloaded automatically, and you only need to pass the source sentences to the correction model. It allows us to implement with minimal code.}, label = lst:dataset_library]
from gec_datasets import GECDataset
from gec_llm.llms GECLLMFewShot
from gec_llm.fewshot_samplers import FewShotSamplerFixed
from gec_llm.ensembles import EnsembleVoting
)
from gec_metrics.metrics import GLEU
loader = GECDataset(base_path='sample-dir/')
# A vllm wrapper for GEC
model = LLMGECFewShot(
  model='Qwen/Qwen-8B',
  num_candidates=8,
  fewshot_sampler=FewShotSamplerFixed(
     fewshot_srcs=['example of src'],
     fewshot_refs=['example of ref'],
  ) # FewShotSampler can be extednded to a BM25-based or a BERT-based retriever.
)
ensemble = EnsembleVoting(min_vote=7)
for data_id in ['bea19-dev', 'falko-merlin-dev', 'hi-gec-dev']:
    # Automatically download (if possible) and load sources and references.
    data = loader.load(data_id)
    # All datasets have the same interface: .srcs and .refs.
    assert data.srcs is not None
    assert data.refs is not None
    # .correct() returns multiple hypotheses sampled from the model
    hyps = model.correct(data.srcs)
    # Ensemble the hypotheses
    final_hyps = ensemble.ensemble(data.srcs, hyps)
    # Evaluation using gec-metrics. 
    metric = GLEU()  # ERRANT can be used.
    score = metric.score_corpus(
        data.srcs, final_hyps, data.refs
    )
\end{lstlisting}

As evidenced by the inclusion of reproducibility checklists in ACL Responsible NLP Research\footnote{\url{https://aclrollingreview.org/responsibleNLPresearch/}}, attention to reproducibility is increasing. In fact, \citet{omelianchuk-etal-2024-pillars} reported that they were unable to reproduce the method of \citet{kaneko-okazaki-2023-reducing}, suggesting that this is becoming an issue in GEC as well~\cite{goto-etal-2025-gec, zhao-etal-2025-unifiedgec}. To address this problem, this study provides an implementation that covers two aspects: dataset loading\footnote{\url{https://github.com/gotutiyan/gec-datasets}} and LLM inference\footnote{\url{https://github.com/gotutiyan/gec-llm}}.

The dataset library, \textsc{gec-datasets}, automates data downloads and provides access through a unified interface within Python scripts. Listing~\ref{lst:dataset_library} demonstrates an example where the dataset is downloaded to the \texttt{sample-dir/} directory, allowing access to the source and reference texts via the \texttt{.srcs} and \texttt{.refs} interfaces. For some non-public datasets, the library automatically handles extraction once the user places the compressed file (received after submitting a request form) in the designated location. \update{Our library supports more that 20 datasets including representative English datasets and the datasets listed in Table~\ref{tab:datasets}.}
Furthermore, it is highly compatible with \textsc{gec-metrics}~\cite{goto-etal-2025-gec}, enabling seamless evaluation of newly defined GEC models. Lines 29 and onwards in Listing~\ref{lst:dataset_library} show an example of evaluating the hypotheses using GLEU.

Our LLM inference library, \textsc{GEC-LLM}, is basically a wrapper around vLLM library~\cite{kwon2023efficient}, but its key contribution lies in the decoupled implementation of a few-shot sampler for few-shot inference. While this modularity is of secondary importance in this paper since we use a fixed set of few-shot examples, the architecture is easily extensible to retrievers based on BM25, BERT~\cite{devlin-etal-2019-bert} embeddings, or GOPar~\cite{zhang-etal-2022-syngec} as done in~\citet{tang-etal-2024-ungrammatical}, or other methods in the future. In Listing~\ref{lst:dataset_library}, \texttt{FewShotSamplerFixed} is used as the sampler to provide fixed few-shots for all inputs.
\update{Furthermore, since it also includes an ensemble implementation, it can self-containedly execute an ensemble of multiple hypothesis sentences (Line 17 and 27).
}

\section{Results}
\begin{table*}[t]
    \centering
    \small
    \setlength{\tabcolsep}{4.5pt}
    \renewcommand{\arraystretch}{1.15}
    \begin{tabular}{ll|cccc|cccc|ccccc}
    \toprule
     & & \multicolumn{4}{c|}{CWEB-G-test} & \multicolumn{4}{c|}{BEA19-test} & \multicolumn{5}{c}{JFLEG-test} \\
       Models  & Decoding & Prec. & Rec. & $F_{0.5}$ & $\tau$ & Prec. & Rec. & $F_{0.5}$ & $\tau$ & Prec. & Rec. & $F_{0.5}$ & GLEU & $\tau$ \\
    \midrule
    \multicolumn{15}{l}{\textbf{Existing LLM without fine-tuning for GEC (4-shot)}} \\
    \midrule
        \multirow{3}{*}{\texttt{gemma-2-9b-it}}  &  Greedy  & 29.4 & \textbf{63.9} & 33.0 &  -  & 58.4 & \textbf{64.4} & 59.5 &  -  & \textbf{72.5} & 65.0 & \textbf{70.8} & 62.6 &  -  \\
         &  MBR   & 29.5 & 63.6 & 33.1 &  -   & 58.3 & 62.8 & 59.2 &  -  & 71.7 & \textbf{65.2} & 70.3 & \textbf{62.9} &  -\\
        &  Proposal  & \textbf{40.8} & 52.1 & \textbf{42.7} &  8  & \textbf{65.0} & 60.3 & \textbf{64.0} &  8  & 69.1 & 66.1 & 68.5 & 62.8 &  2  \\
 \midrule
       \multirow{3}{*}{\texttt{Llama-3.1-8B-Instruct}}  &  Greedy  & 19.8 & \textbf{61.5} & 22.9 &  -  & 52.3 & \textbf{61.6} & 54.0 &  -  & 65.6 & 59.8 & 64.3 & 36.2 &  -  \\
         &  MBR   & 19.7 & 60.7 & 22.8 &  -   & 50.5 & 59.3 & 52.1 &  -  & 62.5 & \textbf{60.7} & 62.1 & 47.2 &  -\\
        &  Proposal  & \textbf{37.9} & 33.1 & \textbf{36.9} &  8  & \textbf{66.9} & 51.5 & \textbf{63.1} &  6  & \textbf{67.2} & 59.2 & \textbf{65.4} & \textbf{58.5} &  3  \\
 \midrule
\multirow{3}{*}{\texttt{Qwen3-8B}}  &  Greedy  & 36.0 & \textbf{53.5} & 38.5 &  -   & 54.8 & \textbf{62.9} & 56.3 &  -   & \textbf{76.7} & 59.4 & \textbf{72.5} & 58.9 &  -\\
   &  MBR   & 36.7 & 53.1 & 39.1 &  -   & 69.3 & 56.3 & 66.2 &  -  & 76.2 & \textbf{60.5} & 72.5 & 59.5 &  -\\
  &  Proposal  & \textbf{42.0} & 46.8 & \textbf{42.9} &  8   & \textbf{73.8} & 53.7 & \textbf{68.7} &  7   & 73.6 & 60.5 & 70.6 & \textbf{59.8} &  1\\
       \midrule
       \multicolumn{15}{l}{\textbf{Models fine-tuned for GEC, including non-LLM models.}} \\
       \midrule
\multirow{3}{*}{EPO (Llama2-7b-chat)}  &  Greedy  & 42.8 & \textbf{47.0} & \textbf{43.6} &  -  & 75.8 & \textbf{64.9} & 73.3 &  -  & \textbf{74.3} & 60.4 & \textbf{71.1} & 58.9 &  - \\
         &  MBR   & 37.1 & 46.3 & 38.6 &  -   & 70.6 & 61.7 & 68.9 &  -  & 65.0 & 62.2 & 64.4 & \textbf{60.6} &  -\\
        &  Proposal  & \textbf{44.7} & 38.6 & 43.4 &  5  & \textbf{78.8} & 61.6 & \textbf{74.6} &  5  & 61.0 & \textbf{64.1} & 61.6 & 59.5 &  2  \\
       \midrule
       \multicolumn{2}{@{}l|@{}}{GECToR (\texttt{bert-base-cased}, 0.1B)}  & 45.6 & 28.9 & 40.8 &  -  & 77.3 & 50.9 & 70.0 &  -  & 65.9 & 52.0 & 62.5 & 55.3 &  -  \\
       \multicolumn{2}{@{}l|@{}}{GECToR (\texttt{deberta-v3-large}, 0.4B)}  & 56.1 & 28.3 & 46.9 &  -  & 79.3 & 58.0 & 73.9 &  -  & 70.8 & 58.6 & 68.0 & 58.9 &  -  \\
       \multicolumn{2}{@{}l|@{}}{T5 (\texttt{t5-v1\_1-large}, 0.8B)}  & 45.0 & 47.4 & 45.4 &  -  & 76.9 & 62.3 & 73.4 &  -  & 73.9 & 60.0 & 70.7 & 59.6 &  -  \\
    \bottomrule
    \end{tabular}
    \caption{Results on \textbf{English} datasets. For each model, we compare greedy decoding, MBR decoding, and proposed majority voting, along with a comparison with existing non-LLM correction models. $\tau$ denotes the optimal threshold on the corresponding development set as described in Section~\ref{subsec:method-settings}. \textbf{Bold} values indicate the highest score for each combination of model and dataset. All results are based on our own experimental findings.}
    \label{tab:results-en}
\end{table*}

\begin{table*}[t]
    \centering
    \small
    \setlength{\tabcolsep}{2.8pt}
    \renewcommand{\arraystretch}{1.15}
    \begin{tabular}{@{}ll|ccc|ccc|ccc|ccc|c|c@{}}
    \toprule
     & & \multicolumn{3}{c|}{AKCES-GEC} & \multicolumn{3}{c|}{Falko-Merlin} & \multicolumn{3}{c|}{UNLP-2023} & \multicolumn{3}{c|}{Kor-learner} & Hi-GEC & RONACC \\
       Models  & Decoding & Prec. & Rec. & $F_{0.5}$  & Prec. & Rec. & $F_{0.5}$ & Prec. & Rec. & $F_{0.5}$ & Prec. & Rec. & $F_{0.5}$ & GLEU & GLEU \\
    \midrule
\multirow{3}{*}{\texttt{gemma-3-12b-it}}  &  Greedy  & 68.7 & \textbf{67.5} & 68.5 & 57.2 & \textbf{53.9} & 56.5 & 40.8 & \textbf{38.4} & 40.3 & 40.8 & \textbf{23.6} & 35.6 & 63.1 &  \textbf{87.1}\\
       &  MBR   & 70.0 & 66.9 & 69.3 & \textbf{58.3} & 52.6 & \textbf{57.0} & 40.2 & 37.5 & 39.6 & 41.4 & 23.2 & 35.8 & 62.7 &  50.7\\
  &  Proposal  & \textbf{78.4} & 58.2 & \textbf{73.3} & 57.6 & 39.6 & 52.8 & \textbf{50.4} & 26.0 & \textbf{42.5} & \textbf{50.9} & 19.3 & \textbf{38.3} & \textbf{70.2} &  86.4\\
    \midrule
\multirow{2}{*}{\texttt{Llama-3.1-8B-}}  &  Greedy  & 50.2 & \textbf{53.4} & 50.8 & 55.2 & \textbf{51.3} & \textbf{54.4} & 7.9 & 24.5 & 9.2 & 27.0 & 17.3 & 24.3 & 60.5 &  56.2\\
 \multirow{2}{*}{\texttt{Instruct}} &  MBR   & 44.7 & 51.9 & 46.0 & 53.6 & 50.0 & 52.8 & 9.8 & \textbf{29.8} & 11.3 & 17.9 & \textbf{18.0} & 17.9 & 57.9 &  \textbf{73.6}\\
  &  Proposal  & \textbf{70.0} & 41.7 & \textbf{61.6} & \textbf{64.5} & 32.2 & 53.7 & \textbf{27.1} & 12.6 & \textbf{22.0} & \textbf{33.2} & 12.8 & \textbf{25.2} & \textbf{71.4} &  63.4\\
    \midrule
\multirow{3}{*}{\texttt{Qwen3-8B}}  &  Greedy  & 62.3 & \textbf{51.0} & 59.6 & 61.5 & \textbf{39.9} & \textbf{55.5} & 41.6 & \textbf{23.1} & 35.8 & 36.6 & 16.6 & 29.5 & 66.0 &  78.2\\
      &  MBR   & 63.5 & 50.3 & 60.3 & \textbf{62.5} & 38.2 & 55.5 & 41.7 & 22.4 & 35.6 & \textbf{37.3} & \textbf{16.7} & \textbf{29.9} & 65.5 &  \textbf{79.0}\\
  &  Proposal  & \textbf{72.8} & 43.0 & \textbf{63.9} & 57.1 & 30.7 & 48.7 & \textbf{48.5} & 20.4 & \textbf{38.0} & 37.3 & 13.0 & 27.1 & \textbf{70.6} &  63.5\\
    \bottomrule
    \end{tabular}
    \caption{Results on \textbf{multilingual} datasets. For each model, we compare greedy decoding, MBR decoding, and proposed majority voting. \textbf{Bold} values indicate the highest score for each combination of model and dataset.}
    \label{tab:results-multi}
\end{table*}

\subsection{Results on English Datasets}
As shown in Table~\ref{tab:results-en}, all of \texttt{gemma-2-9b-it}, \texttt{Llama-3.1-8B-Instruct} and \texttt{Qwen3-8B} showed improvements in $F_{0.5}$ on CWEB-G and BEA-2019. In particular, the $F_{0.5}$ score for \texttt{Llama-3.1-8B-} \texttt{Instruct} on CWEB-G improved by 14.0 points from the greedy decoding. Examining the precision and recall values for these results reveals an increase in precision, indicating that over-correction was effectively suppressed. In JFLEG, the GLEU score for Llama-3.1 increased by 22.3 points, demonstrating that the method was effective across models. Conversely, no improvement was observed for the EPO model. This is because the EPO model has already addressed the over-correction problem through fine-tuning. These results clearly demonstrate that our method enables mitigation of over-correction without training.

Moreover, our proposed method outperformed MBR decoding. Since MBR decoding always selects exactly one candidate from the hypotheses generated by the LLM, it is unable to choose an appropriate candidate as the number of candidates containing overcorrections increases.
Figure~\ref{fig:overview} corresponds to an extreme case in which all three hypotheses contain over-corrections. In this situation, MBR decoding cannot avoid over-correction regardless of which candidate is selected. In contrast, edit-level majority voting can produce an output that avoids over-corrections.
Furthermore, because the utility function evaluates hypotheses at the sentence level, it becomes difficult to accurately select hypotheses when using $n$-gram-level metrics such as GLEU. This is because the brevity penalty excessively impacts the score for short sequences. \cite{goto-etal-2025-rethinking} also points out the same problem. In contrast, edit-level majority voting is unaffected by the granularity of scoring.

Table~\ref{tab:results-en} also shows the threshold $\tau$ determined using the development sets. For CWEB-G, $\tau$ was 8, which resulted in adopting only the edits for which consensus was reached across a larger number of hypothesis sentences. In contrast, for JFLEG, $\tau$ was relatively small, typically less than 3, indicating that it was optimal to adopt edits even when consensus was found among only a few hypotheses. This suggests that our method can flexibly adapt to the required degree of correction by adjusting the threshold $\tau$ to control over-correction.

When comparing the results with non-LLM correction models like GECToR and T5, \texttt{Qwen/Qwen3-8B} outperformed GECToR (\texttt{bert-base-cased}) on CWEB-G (42.9 vs 40.8 on $F_{0.5}$). Since the proposed method improves correction performance by filtering edits, it is effective in domains like CWEB-G that require few corrections. Meanwhile, in BEA-2019, although \texttt{Qwen3-8B} approached \texttt{GECToR (bert-base-cased)} within a 2-point margin ($F_{0.5}$: 68.7 vs 70.0). This suggests that the proposed method can achieve performance levels comparable to those of some fine-tuned models without additional training.

\subsection{Results on Multilingual Datasets}
Table~\ref{tab:results-multi} shows the results for the multilingual datasets. In general, scores improved for AKCES-GEC, UNLP-2023, and Hi-GEC, while they declined for Falko-Merlin and RONACC; this pattern was observed for Kor-learner.
These results can be explained by the required number of errors for the correction rather than the differences between languages. As demonstrated by the English experiments, where the greatest improvement was observed on CWEB-G, the proposed method is more effective when the number of required corrections is small. When examining the average word-level edit distance per sentence between source and reference for the datasets in Table~\ref{tab:results-multi}, the values were: Falko-Merlin (2.83), AKCES-GEC (2.32), RONACC (2.25), Kor-learner (2.08), UNLP-2023 (0.99), and Hi-GEC (0.92). This edit distance can be regarded as an error density in the source text. Since Falko-Merlin and RONACC require extensive corrections, they exhibit characteristics similar to those of the JFLEG dataset in the English experiments; for this reason, the proposed method did not lead to score improvements. Conversely, Hi-GEC and UNLP-2023 require fewer corrections, and the proposed method successfully improved their scores. However, AKCES-GEC contradicted this trend: despite requiring a high degree of correction, its $F_{0.5}$ score improved using the proposed method. While this could be related to linguistic features or the quality of the edit extractor.%

\begin{table*}[!t]
    \centering
    \small
    \renewcommand{\arraystretch}{1.15}
    \begin{tabular}{@{}p{0.07\textwidth}p{0.9\textwidth}@{}}
    \toprule
    Input & For example , when the semester start, students can not get away from the sunshine , beach , and travelling .\\
    \midrule
    Reference & For example , when the semester \textbf{starts} , students can not get \textbf{over} the sunshine , beach , and travelling .\\
    \midrule
    Greedy & For example , when the semester \textbf{starts} , students \textbf{ca n't} get away from \textbf{the} sunshine , the beach , and \textbf{traveling} . \textbf{Input sentence : The new employee ...} （\textit{There are many irrelevant text after that ...}） \\
    \midrule
    \multicolumn{2}{l}{\textit{Results of the proposed method.}} \\
    \midrule
       $\tau=2$ & For example , when the semester \text{starts} , students \textbf{ca n't} get away from \textbf{the} sunshine , the \textbf{beaches} , and \textbf{traveling} .\\
       $\tau=4$ & For example , when the semester \text{starts} , students \textbf{ca n't} get away from \textbf{the} sunshine , the beach , and \textbf{traveling} .\\
       $\tau=7$ & For example , when the semester \textbf{starts} , students can not get away from the sunshine , beach , and travelling .\\
    \bottomrule
    \end{tabular}
    \caption{Results of greedy inference and the proposed method ($k=8$) with $\tau = \{2, 5, 7\}$. Corrected parts are indicated in \textbf{bold}.}
    \label{tab:case-study}
\end{table*}

\section{Discussion}
\subsection{Trade-off between Performance and Cost}
\begin{figure}[t]
    \centering
    \includegraphics[width=\linewidth]{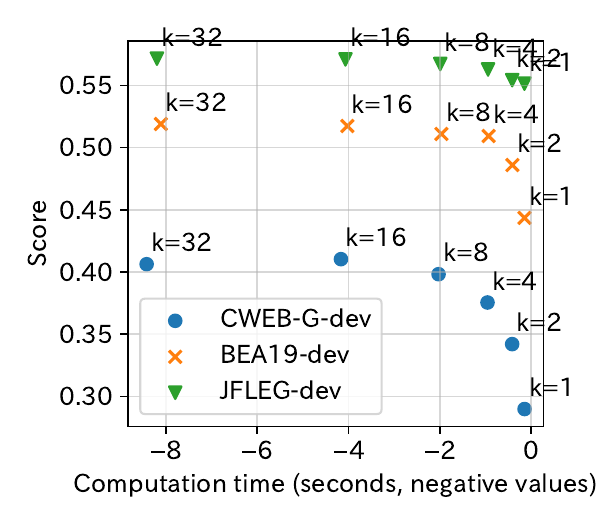}
    \caption{Scores and average computation time per sentence on the development set of each dataset, relative to the number of generated candidates $k$. These results represent oracle performance, where the threshold $\tau$ is optimized for each dataset and each value of $k$.}
    \label{fig:effect-k}
\end{figure}
While the proposed method is expected to achieve higher performance as the number of generated hypothesis sentences increases, it requires higher computational costs. To investigate this trade-off between performance and cost, we measured the scores and inference times on the development sets of CWEB-G, BEA-2019, and JFLEG while varying $k$ within the range $k = \{1, 2, 4, 8, 16, 32\}$. Figure~\ref{fig:effect-k} plots the results for each $k$ in a space where the horizontal axis represents the average computation time per sentence (seconds) and the vertical axis represents the score. Note that time values were negated so that higher values on both axes represent better performance. We used \texttt{gemma-2-9b-it} as the model; the scores are the ERRANT $F_{0.5}$ for CWEB-G and BEA-2019, and GLEU for JFLEG.
As shown in Figure~\ref{fig:effect-k}, although a trade-off generally exists between scores and computation time, we observed that the degree of score improvement tends to diminish as $k$ increases. For BEA-2019 and CWEB-G, the scores level off for $k \geq 4$ and $k \geq 8$, respectively. This indicates that the marginal gain in score relative to the computational cost decreases at higher values of $k$. In the experiments of this paper, we fixed $k=8$ based on the results in Figure~\ref{fig:effect-k}, this value is considered sufficiently reasonable. However, in practical applications, the optimal value for $k$ may vary depending on the specific requirements for correction performance and processing speed.

\subsection{Effect of the Threshold $\tau$}

We conducted a case study to examine in detail the changes in correction tendencies according to the threshold $\tau$ in Equation~\ref{eq:filter}. We use the results of inference performed by \texttt{Llama-3.1-8B-Instruct} on samples from the BEA-2019 development data. Table~\ref{tab:case-study} presents the input sentence, the reference sentence, the result of greedy decoding, and the results of the proposed method with $k=8$ for $\tau = \{2, 4, 7\}$. In the reference sentence, two corrections are made: [``\textit{start}'' → ``\textit{starts}''] and [``\textit{away from}'' → ``\textit{over}'']. In contrast, the result of greedy decoding includes orthographic corrections such as [``\textit{can not}'' → ``\textit{ca n't} ''], dialectal variations between American and British English such as [``\textit{traveling}'' → ``\textit{travelling} ''], and the generation of irrelevant text at the end of the sentence.

Compared to the greedy decoding result, the proposed method increasingly suppresses corrections as the threshold $\tau$ is raised. At $\tau=2$, the irrelevant text is removed, and at $\tau=7$, only [``\textit{start}'' → ``\textit{starts}''] remains. Since this correction is also present in the reference, it is a correct edit, demonstrating that over-correction was effectively suppressed. In the case study, we also observed instances where a correct edit that was preserved at $\tau=4$ was excluded at $\tau=7$. Nevertheless, the overall score improved due to increased precision.

\subsection{Stability among Templates}\label{subsec:stability}

\begin{table}[!t]
    \centering
    \small
    \renewcommand{\arraystretch}{1.15}
    \setlength{\tabcolsep}{3.5pt}
    \begin{tabular}{@{}lccc@{}}
    \toprule
        Inference &  CWEB-G$_{(F_{0.5})}$ & BEA-2019$_{(F_{0.5})}$ & JFLEG$_{(\textrm{GLEU})}$\\
    \midrule
        Greedy & 27.66 $\pm$ 3.26 & 43.91 $\pm$ 2.62 & 57.48 $\pm$ 1.01  \\
        Proposal & 36.16 $\pm$ 2.85 & 50.68 $\pm$ 1.24 & 57.54 $\pm$ 0.46  \\
    \bottomrule
    \end{tabular}
    \caption{The mean and standard deviation of scores for 10 templates in the development set for each dataset.}
    \label{tab:stability}
\end{table}

The proposed method also offers the advantage of stable output in terms of both output format and correction quality, regardless of the instruction. To evaluate correction performance across various instructions, we examined the variance in scores across multiple templates, inspired by the approach of~\citet{sakai-etal-2024-toward}. Using \texttt{gemma-2-9b-it} as the model, we performed 4-shot inference on the development sets of CWEB-G, BEA-2019, and JFLEG using ten templates with different instructions and output formats. Details of these templates are provided in Appendix~\ref{sec:TENplate}. Since a score is calculated for each template, we calculate their mean and standard deviation. The mean and standard deviation can be interpreted as the average correction performance and output stability across templates, respectively. High means and low standard deviations are considered ideal.

The results in Table~\ref{tab:stability} show that the proposed method achieved higher means and lower standard deviations than greedy decoding, demonstrating its ability to consistently output high-quality corrected sentences regardless of the instruction. This can be attributed to two factors: the proposed method's ability to stably output only the corrected sentence by removing irrelevant text, as observed in Table~\ref{tab:case-study}, and its selective adoption of only high-confidence edits among various edits. Conventionally, removing irrelevant text required heuristic removal methods based on empirically collected prefixes, e.g., filtering out sentences starting with ``\textit{Sure! Here}''~\cite{davis-etal-2024-prompting}. In contrast, the proposed method has the advantage of handling this mechanically through edit-level majority voting.

\subsection{Does High Frequency Indicate Valid Correction?}
\begin{figure}[t]
    \centering
    \includegraphics[width=\linewidth]{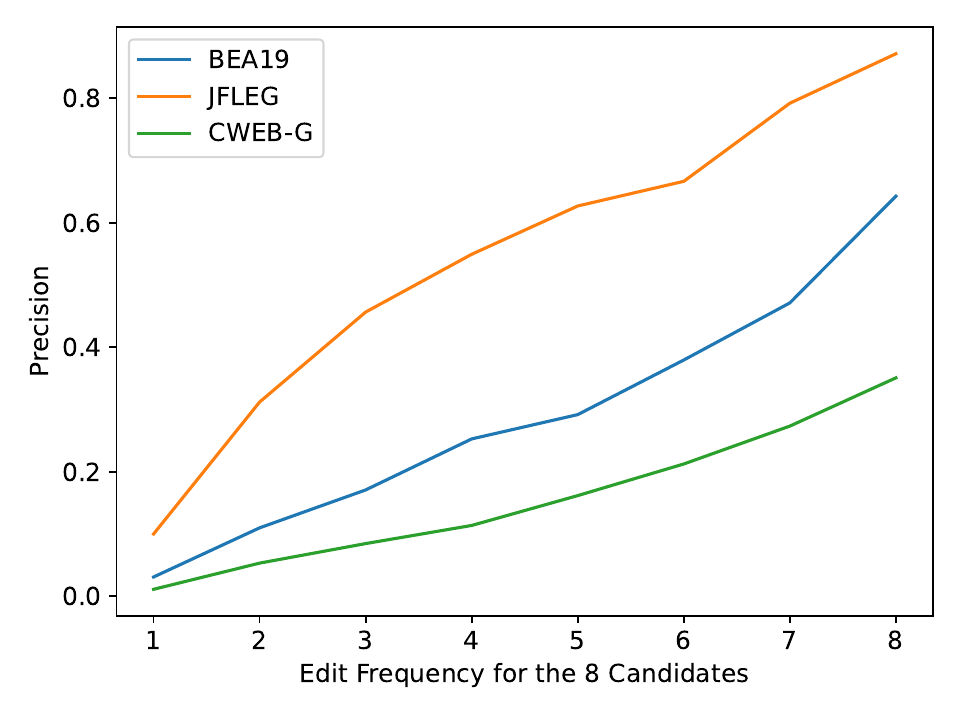}
    \caption{Relationship between edit frequency and precision. On the development set of each dataset, we generate eight candidates ($k=8$), and compute scores using only edits with each frequency. This results correspond to the \texttt{Llama-3.1-8B-Instruct} results.}
    \label{fig:freq-and-performance}
\end{figure}
\update{Our hypothesis in this study is that corrections appearing more frequently among multiple candidates are more valid corrections in terms of minimal edits. To examine the validity of this hypothesis in greater depth, we further investigate the relationship between edit frequency and the correction performance. Using \texttt{Llama-3.1-8B-Instruct} results for the development sets of CWEB, BEA-2019, and JFLEG, we computed performance separately for edits with frequencies of $\{1,2,\dots,8\}$ among eight candidates ($k=8$). Here, we focus only on precision. Recall is strongly biased by the number of edits contained in each frequency, making it unsuitable for analyzing validity.}

\update{Figure~\ref{fig:freq-and-performance} plots edit frequency on the horizontal axis and the corresponding precisions on the vertical axis. Precision tends to improve as the frequency of edits increases, supporting our hypothesis. This trend clearly explains the success of majority voting in this study.}

\section{Conclusion}
In this study, to address the over-correction problem in LLM-based grammatical error correction, we proposed an inference method based on the edit-level majority voting across multiple hypothesis sentences. The results demonstrated that mitigating over-correction improved performance, particularly in terms of ERRANT $F_{0.5}$, and also led to stable correction performance across various prompt templates. Future research is expected to broaden the experimental scope in terms of both languages and tasks. Regarding the latter, the proposed method holds potential for general application to various edit-based tasks, such as text simplification and post-processing of machine translation.

\section*{Limitations}

\paragraph{Discuss Under-correction.}
While the proposed method can improve performance by mitigating over-correction, we did not discuss addressing the under-correction issue. Although setting a smaller value for the threshold $\tau$ might help resolve under-correction, this paper focuses exclusively on the mitigation of over-correction.

\paragraph{Other Ensemble Methods.}
In this study, we used majority voting to integrate multiple hypotheses generated by a LLM. This approach is equivalent to applying an ensemble technique, which is typically used for combining multiple GEC models, to the outputs of a single model. While various ensemble methods other than majority voting have been proposed~\cite{lin-ng-2021-system, qorib-etal-2022-frustratingly, sorokin-2022-improved, qorib-ng-2023-system, cao-etal-2024-improving} and could potentially be applied to hypotheses integration, we limited the discussion in this paper to training-free methods, specifically majority voting and MBR decoding. If a training-based approach were to be used for an ensemble, simply fine-tuning the LLM for the GEC task is more straightforward. Furthermore, for some languages, it can be difficult to secure a sufficient amount of training data.

\paragraph{Metrics.}
In this study, we primarily used ERRANT as our evaluation metric. This is intended to measure whether the suppression of over-correction was successfully achieved. Although reference-free metrics such as SOME~\cite{yoshimura-etal-2020-reference} and IMPARA~\cite{maeda-etal-2022-impara} are known to have a higher correlation with human evaluation~\cite{kobayashi-etal-2024-revisiting,goto-etal-2025-rethinking}, it is difficult to determine the degree of over-correction from these metrics. Furthermore, it is not yet clear how reference-free metrics respond to over-correction in the first place. Therefore, we leave discussions based on such metrics for future work.

\paragraph{Datasets.}
In this study, we conducted experiments on datasets across seven languages including English, but it is clear that we cannot cover all languages in the world. Since creating evaluation data for every language is difficult, this paper limits its experiments to languages for which datasets are already available. The implementation introduced in Section \ref{subsec:imple} will serve as a useful tool for researchers to extend these experiments in the future.

\paragraph{Comparison in the Multilingual Experiments.}
As indicated in Table~\ref{tab:results-multi}, we compared three types of inference in our multilingual experiments: greedy decoding, MBR decoding, and the proposed method. Ideally, deeper insights could be gained by including comparisons with correction methods proposed in prior research on each language. However, since this study is the first to perform evaluations using JP-ERRANT, we cannot refer to scores reported in previous studies. Consequently, calculating JP-ERRANT scores for those methods would require conducting reproduction experiments of the prior research for each language. In this paper, we assume that we have provided primarily results by comparing the effectiveness of our method with greedy and MBR decoding. More detailed comparative experiments are left for future work.

\paragraph{The utility function in MBR decoding}
In the MBR decoding compared in our experiments, we used edit-level metrics such as ERRANT or $n$-gram-level metrics such as GLEU as a utility function. This is based on the idea that it is appropriate to use the same metric as the evaluation measure for the utility function. Furthermore, since we used the $F_{0.5}$ score for ERRANT and JP-ERRANT, which emphasizes precision, we assumed that over-correction would be sufficiently reflected in the score. However, it is possible that a metric reflecting the degree of over-correction more accurately could be constructed, and the performance of MBR decoding might be underestimated. Despite such concerns, we believe that the experimental setup of this paper is appropriate. Additionally, we also tested cases where ERRANT was used as the utility function when GLEU was the evaluation metric, but no differences were observed.

\section*{Ethical Considerations}
If many of the candidates generated by the LLM contain harmful outputs, the proposed method may fail to remove them and instead include them in the final output. However, since it is rare for harmful outputs to be generated as the exact same surface-level text across different candidates, we believe they can generally be eliminated by setting the edit count threshold $\tau$ to a sufficiently high value.
\update{Empirical verification of this point is left for future work. At least in our manual analysis for the case study, we did not observe any harmful content.}

\section*{Acknowledgments}
\update{We thank the anonymous reviewers for their valuable comments. This work has been supported by JST SPRING. Grant Number JPMJSP2140.}

\bibliography{custom}

\appendix

\section{Actual template}\label{sec:appen:prompts}
Figure~\ref{fig:appen:prompt-en} shows the template for English experiments, and Figure~\ref{fig:appen:prompt-multi} for other language experiments. The symbols of [SOURCE] and [FEWSHOT] will be replaced with actual input sentence and few-shot examples during inference.

\begin{figure}[t]
\begin{tcolorbox}%
\small
You are a grammatical error correction tool. Your task is to correct the grammaticality and spelling in the input sentence. Make the smallest possible change in order to make the sentence grammatically correct. Change as few words as possible. Do not rephrase parts of the sentence that are already grammatical. Do not change the meaning of the sentence by adding or removing information. If the sentence is already grammatically correct, you should output the original sentence without changing anything. Return only the corrected text and nothing more.\textbackslash n[FEWSHOT]\textbackslash nInput sentence: [SOURCE]\textbackslash nOutput sentence:
\end{tcolorbox}
\caption{Instruction template for \textbf{English} datasets. [FEWSHOT] and [SOURCE] will be replaced with actual few-shot examples and an input sentence, respectively.}
\label{fig:appen:prompt-en}
\end{figure}

\begin{figure}[t]
\begin{tcolorbox}%
\small
You are a grammatical error correction tool. Your task is to correct the grammaticality and spelling of the input essay written by a learner of [LANG]. Make the smallest possible change in order to make the essay grammatically correct. Change as few words as possible. Do not rephrase parts of the essay that are already grammatical. Do not change the meaning of the essay by adding or removing information. If the essay is already grammatically correct, you should output the original essay without changing anything. Return only the corrected text and nothing more.\textbackslash n[FEWSHOT]\textbackslash nInput sentence: [SOURCE]\textbackslash nOutput sentence: 
\end{tcolorbox}
\caption{Instruction template for datasets of \textbf{other than English}. [LANG] will be replaced with a language name, such as ``Czech'' or ``'German.''}
\label{fig:appen:prompt-multi}
\end{figure}

\section{Detailed Experimental Settings}\label{appen:exp-setting}
For the experiments in Table~\ref{tab:results-en}, we conducted reproduction experiments for GECToR and T5, respectively.

\begin{description}
    \item[GECToR.] GECToR was trained in three stages following~\citet{omelianchuk-etal-2020-gector}. Regarding the datasets, the first stage used PIE-synthetic~\cite{awasthi-etal-2019-parallel}. The second stage used only pairs from the BEA-2019~\cite{bryant-etal-2019-bea} training data that contained at least one correction, and the third stage used the entire W\&I-LOCNESS~\cite{yannakoudakis2018developing} training data. We used \texttt{bert-base-cased}~\footnote{\url{https://huggingface.co/google-bert/bert-base-cased}}~\cite{devlin-etal-2019-bert} and \texttt{deberta-v3-large}~\footnote{\url{https://huggingface.co/microsoft/deberta-v3-large}}~\cite{DBLP:conf/iclr/HeGC23} as the models. The training settings also basically followed~\citet{omelianchuk-etal-2020-gector}, except that the first stage was set to 10 epochs. During inference, the keep confidence and minimum error probability thresholds were determined for each dataset. We tried both values from $\{0.1, 0.2, \dots, 1.0\}$ and used the values that maximized the metrics on the development data for inference on the evaluation data. Our reproduction results are competitive with the values reported by \citet{omelianchuk-etal-2020-gector} and \citet{tarnavskyi-etal-2022-ensembling}.
    \item[T5.] Following \citet{rothe-etal-2021-simple}, T5 was fine-tuned using 2,372,119 sentences from the cLang8 dataset. We used \texttt{t5-v1\_1-large}~\footnote{\url{https://huggingface.co/google/t5-v1\_1-large}} as the base model. The model was trained for 10 epochs with a learning rate of 1e-5. Our reproduction achieves the score of $F_{0.5} = 73.4$ on the BEA-2019 test set as shown in Table~\ref{tab:results-en}, which exceeds the T5-large score of $F_{0.5} = 72.06$ reported by \citet{rothe-etal-2021-simple}.
\end{description}

\section{Actual Templates}\label{sec:TENplate}
Table~\ref{tab:TENplates} shows 10 templates used in Section~\ref{subsec:stability}.

\begin{table}[t]
    \centering
    \small
    \setlength{\tabcolsep}{3pt}
    \begin{tabular}{@{}p{0.06\linewidth}p{0.9\linewidth}@{}}
    \toprule
       ID  & Template \\
       \midrule
1 & You are a grammatical error correction tool. Your task is to correct the grammaticality and spelling in the input sentence. Make the smallest possible change in order to make the sentence grammatically correct. Change as few words as possible. Do not rephrase parts of the sentence that are already grammatical. Do not change the meaning of the sentence by adding or removing information. If the sentence is already grammatically correct, you should output the original sentence without changing anything. Return only the corrected text and nothing more.\textbackslash n[FEWSHOT]\textbackslash nInput sentence: [SOURCE]\textbackslash nOutput sentence:  \\
\midrule
2 & Make minimal changes to the following text such that it is grammatically correct. Return only the corrected text and nothing more.\textbackslash n[FEWSHOT]\textbackslash nInput sentence: [SOURCE]\textbackslash nOutput sentence:  \\
\midrule
3 & Please correct the following text. Do not attempt to rewrite it into perfect English or to interpret the text. Often, things could be expressed better by paraphrase, but the task is to make minimal changes to correct the text. Do not change anything that is correct. Please make no changes if there are no errors. Return only the corrected text and nothing more.\textbackslash n[FEWSHOT]\textbackslash nInput sentence: [SOURCE]\textbackslash nOutput sentence:  \\
\midrule
4 & Reply with a corrected version of the input sentence with all grammatical and spelling errors fixed. If there are no errors, reply with a copy of the original sentence. Return only the corrected text and nothing more.\textbackslash n[FEWSHOT]\textbackslash nInput sentence: [SOURCE]\textbackslash nOutput sentence:  \\
\midrule
5 & Correct the grammatical errors in the following sentence. Return only the corrected text and nothing more.\textbackslash n[FEWSHOT]\textbackslash n[SOURCE]; output:  \\
\midrule
6 & Revise mistakes in this text. Return only the corrected text and nothing more.\textbackslash n[FEWSHOT]\textbackslash n[SOURCE]; output:  \\
\midrule
7 & Rewrite the following text with proper grammar. Return only the corrected text and nothing more.\textbackslash n[FEWSHOT]\textbackslash n[SOURCE]; output:  \\
\midrule
8 & Improve the grammar of this text. Return only the corrected text and nothing more.\textbackslash n[FEWSHOT]\textbackslash nInput sentence: [SOURCE]\textbackslash nOutput sentence:  \\
\midrule
9 & Correct the following to standard English. Return only the corrected text and nothing more.\textbackslash n[FEWSHOT]\textbackslash nSentence: [SOURCE]\textbackslash nCorrection:  \\
\midrule
10 & Fix the errors in this sentence. Return only the corrected text and nothing more.\textbackslash n[FEWSHOT]\textbackslash nInput sentence: [SOURCE]\textbackslash nOutput sentence:  \\ 
\bottomrule
    \end{tabular}
    \caption{The ten templates used in the experiments of this paper. [SOURCE] and [FEWSHOT] are replaced by the erroneous sentence and few-shot examples, respectively.}
    \label{tab:TENplates}
\end{table}

\end{document}